\documentclass[a4paper]{svproc}
\usepackage{url}
\usepackage{graphicx}

\usepackage{amsmath}
\usepackage{amssymb}
\usepackage{enumitem}
\usepackage{siunitx}
\usepackage{float}
\usepackage[symbol]{footmisc}
\usepackage{hyperref}

\newcommand{\RNum}[1]{\lowercase\expandafter{\romannumeral #1\relax}}
\newcommand{\RNumb}[1]{\uppercase\expandafter{\romannumeral #1\relax}}

\begin{document}
\mainmatter              
\title{Flying Hydraulically Amplified Electrostatic Gripper System for Aerial Object Manipulation}
\titlerunning{Flying Electrostatic Gripper}  
%
\author{Dario Tscholl \and Stephan-Daniel Gravert \and Aurel X. Appius \and Robert K. Katzschmann}
\authorrunning{Tscholl et al.} 
%
\tocauthor{Ivar Ekeland, Roger Temam, Jeffrey Dean, David Grove,
Craig Chambers, Kim B. Bruce, and Elisa Bertino}
\institute{Soft Robotics Lab,\\ ETH Zurich,\\ Tannenstrasse 3, 8092 Zurich, CH\\
\email{rkk@ethz.ch}\\ 
\texttt{https://srl.ethz.ch/}}

\maketitle              

\begin{abstract}
Rapid and versatile object manipulation in air is an open challenge. An energy-efficient and adaptive soft gripper combined with an agile aerial vehicle could revolutionize aerial robotic manipulation in areas such as warehousing. 
This paper presents a bio-inspired gripper powered by hydraulically amplified electrostatic actuators mounted to a quadcopter that can interact safely and naturally with its environment. Our gripping concept is motivated by an eagle's foot. Our custom multi-actuator concept is inspired by a scorpion tail design (consisting of a base electrode with pouches stacked adjacently) and spider-inspired joints (classic pouch motors with a flexible hinge layer). A hybrid of these two designs realizes a higher force output under moderate deflections of up to 25$^\circ$ compared to single-hinge concepts. In addition, sandwiching the hinge layer improves the robustness of the gripper. 
For the first time, we show that soft manipulation in air is possible using electrostatic actuation. This study demonstrates the potential of untethered hydraulically amplified actuators in aerial robotic manipulation. Our proof of concept opens up the use of hydraulic electrostatic actuators in mobile aerial systems. \footnote[2]{\url{https://youtube.com/watch?v=7PmZ8C0Ji08}} \footnote[4]{This is a preprint of the following chapter: Tscholl et al., Flying Hydraulically Amplified Electrostatic Gripper System for Aerial Object Manipulation, published in Robotics Research, edited by Aude Billard, Tamim Asfour \& Oussama Khatib, 2023, Springer reproduced with permission of Springer. The final authenticated version is available online at: \url{http://dx.doi.org/10.1007/978-3-031-25555-7\_25}}
\end{abstract}
\newpage
\section{Introduction}
    Soft robotics is a sub-field of robotics that nowadays enjoys rapidly growing interest. This research area works with highly compliant materials similar to those found in living organisms. Working with soft materials enables engineers to build technology that mimics the design principles found in nature more thoroughly. Furthermore, soft materials make systems more capable of complying with many different tasks while making them more user-friendly by allowing continuous deformation. This concept is particularly interesting for applications that require the interaction of systems with their environment. A continually growing and substantial market for robotic automation is found in warehousing. While aerial delivery drones capture headlines, the pace of adoption of drones in warehouses has shown great acceleration. Warehousing constitutes 30\% of the cost of logistics in the US. The rise of e-commerce, greater customer service demands of retail stores and a shortage of skilled labor intensify competition for efficient warehouse operations \cite{warehousedrone}. A few companies like Soft Robotics Inc. already incorporate soft robotic technology in their automated production facilities. Soft robotic technology enables manipulating traditionally hard-to-grasp objects while being compliant with many different sized and shaped bodies. Combining the advantage of soft robotic systems with aerial vehicles would make for a revolutionary approach to warehouse manipulation. We choose to use hydraulically amplified electrostatic actuators, also referred to as HASEL actuators, because they enable muscle-like performance while being lightweight and providing fast actuation speeds. Especially in fields where mass is critical, like in aeronautics, those properties become attractive. Since electrostatic actuators have been used almost exclusively in a laboratory environment, this paper aims to demonstrate the potential of HASELs as part of a more complex system to eventually tackle real industry needs. 
    
    \subsection{Objectives}
\begin{figure}[htp]
        \centering
        \includegraphics[width=\columnwidth] {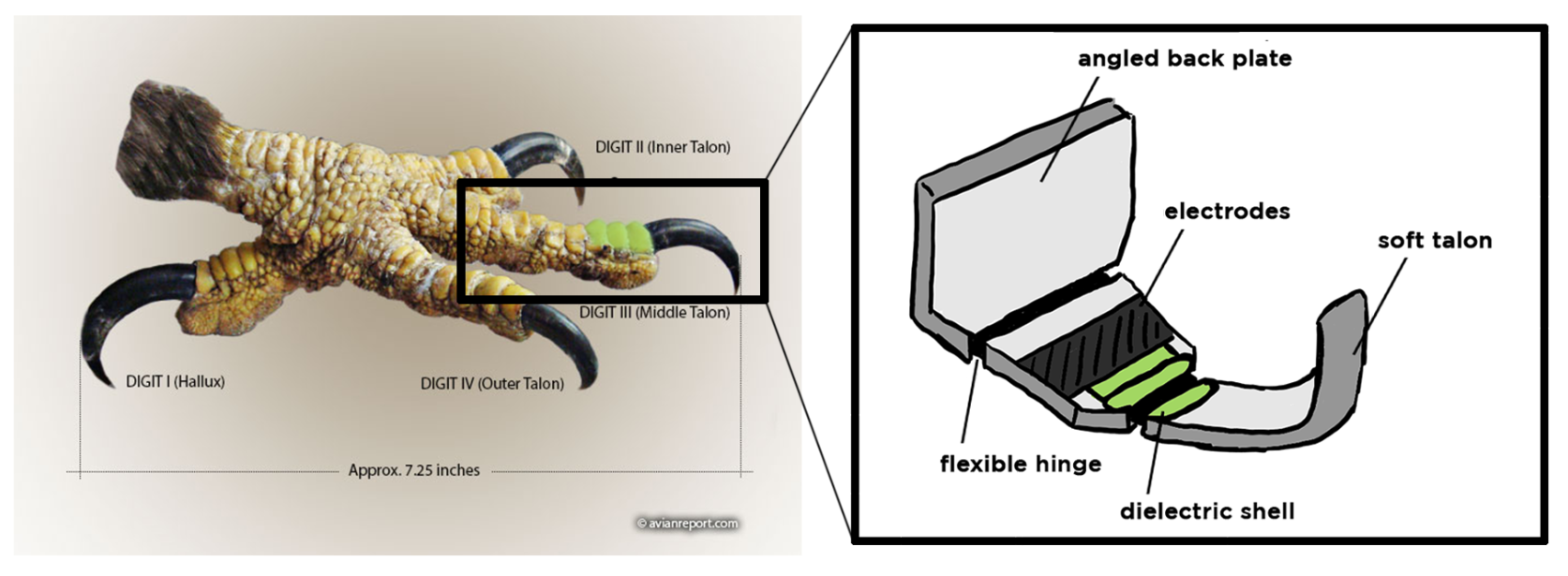}
        \caption{Picture of an eagle's foot on the left as taken from \cite{Eagle_Foot_Pic}. Early visual of a potential implementation based on a bio-inspired design on the right. The main area for actuating the talon is highlighted in green and can be found close to the talon. The idea for the actuator was to narrow the area of rotation and place it as far to the front as possible.}
        \label{fig:foot_eagle}
    \end{figure}

    The main objective of this paper was to design and manufacture a soft robotic gripper for aerial object manipulation using HASEL actuators. As a flying platform, we used the RAPTOR quadcopter developed in parallel and explained in more detail in Section \ref{RAPTOR_quadcopter}. The goal of RAPTOR is dynamic picking and placing of objects in a swooping motion, which is very similar to an eagle's way of hunting. Thus, for the development of this gripper, we investigated the functionality of an eagle's foot. When eagles hunt their prey, they mainly actuate the very front of their toes and barely displace their tarsus. The gripping motion comes primarily from the rotation of the talons. A preliminary illustration of the envisioned concept can be seen in Figure \ref{fig:foot_eagle}. The actual strength of an eagle's grip stems from his upper leg. As shown in \cite{prosthetichand}, a bio-inspired, tendon-driven design would not be feasible for drone integration due to its large size. Consequently, a design inspired by \cite{spider} was sought after.
    \subsection{Related Work} \label{related work}
    \subsubsection{Soft Robotics on Aerial Vehicles.}
    The idea of implementing a gripper on a drone is, in principal, not a novelty \cite{HardFlying_Gripper}. Nonetheless, pioneering designs were rather large and made using classical rigid robotics. Furthermore, the capacity was minimal, only allowing to lift objects weighing 58 g. With time, the force output of newly proposed drone grippers increased substantially, now being able to exceed forces up to 30.6 N \cite{HardFlying_GripperStrong}. Recently, soft robotics has occurred more often in aerial vehicles. The most common examples feature tendon-driven grippers that focus on aggressive and dynamic grasping \cite{SoftFlying_Gripper,Eagle_Gripper}. The issue with tendon-driven actuators is that they still require rigid motors like a Servo motor and thus, do not offer the compliance benefit of soft robotics. One of the most common types of soft robotics are fluidic actuators. Drone grippers of this type already exist and can even demonstrate impressive perching capabilities \cite{PenumaticFlying_Gripper}. However, fluidic actuators struggle with a trade-off between high actuation strain/pressure (determining the force and actuation frequency) and payload capacity. Hence, they are generally more favorable in static applications.
    \subsubsection{Hydraulically amplified self-healing electrostatic actuator}
    Over the last few years a new actuator type called the hydraulically-amplified self-healing electrostatic actuator has proven to show potential in many different applications \cite{haselrothemund}. HASELs have high power density and are efficient when driven by a recuperating high voltage power supply. They are able to exert high forces \cite{haselcharacteristicl} and specific torque outperforming comparable electrical servo motors \cite{spider}. Certain variations even exceed muscle-like performance in specific areas, showing up to 24\% of strain \cite{Side_Electrodes}. Furthermore, HASELs offer fast actuation speeds and low power consumption that surpasses a DC motor \cite{prosthetichand}, making them well-suited for use in soft robotic systems.
    \subsubsection{HASEL Gripper Concepts.}
    The most straightforward approach for designing a HASEL gripper is to stack single Donut HASELs and have at least two stacks facing opposite to each other \cite{donutpapbase}. Changing the pouch from a dimple to a quadrant design allows for additional improvement in the overall strain performance when dealing with larger stacks \cite{haselimproved}. Over time, new HASEL actuator concepts, such as planar HASELs, surfaced, leading to novel, more dynamic gripper designs. Unlike in the Donut HASELs, the electrodes of planar HASEL grippers are located at the end. A backbone plate or strain limiting layer is added to confine the extension of the dielectric shell in one direction, leading to a unidirectional bending motion \cite{haselimproved,simplegrip}. 

    Applying electrodes on the sides instead of the ends results in a so-called High Strain-Peano-HASEL, a variation of a classical Peano-HASEL but exhibiting almost 40\% more strain at low forces \cite{Side_Electrodes}. While allowing for a much denser design length-wise, the large width of the actuation unit is not suitable for an eagle-like gripper requiring a compact design.

    Current research focuses on improving the classic pouch motors \cite{pouchmotor} to achieve high specific torque outputs in combination with a lightweight design. Making use of bio-mimetic elements, certain HASEL's have the ability to exceed a torque output of a servo motor with similar dimensions \cite{spider}. Showing an increase in strength by a factor of five compared to previous HASEL designs, this advancement poses a big step towards a potential use in more complex systems.

    Another gripper concept where the force is induced outside of the joints is found in the field of prosthetics. There, Peano-HASELs behave like artificial muscles in combination with tendons acting as ligaments \cite{prosthetichand}. Despite the higher force output of Peano-HASELs \cite{peanohasel}, outweighing DC motors in terms of speed and energy consumption, the pinching force is considerably smaller. Moreover, the size requirement imposes a notable constraint on space limitations, preventing its use in aeronautical applications.
\section{Methods}
In this section, we compare different soft actuator types, hence not considering systems utilizing traditional electrical motors. Furthermore, we elaborate on the quadcopter we used. The physics and challenges of working with hydraulically amplified electrostatic actuators will be discussed briefly and followed by an insight into our manufacturing process.
    \subsection{Actuator Type Comparison}
    \subsubsection{Fluidic Actuators.} In soft robotics, fluidic actuators are predominantly used because of their high force output \cite{PracticalApproach}. Especially for lightweight, compliant manipulators and gripping systems, but also in crawling and jumping machines moved by expanding gas chambers. However, fluidic actuators require a supply of pressurized gas or liquid \cite{donutpapbase}. Additionally, a network consisting of channels, tubes, and pumps is needed, leading to added weight and losses in speed and efficiency due to fluid transportation.
    \subsubsection{Dielectric Elastomer Actuators.} Another actuation type, such as dielectric elastomer actuators (DEAs), do offer high efficiency (80\%), high actuation strain ($>$100\%) and are self-sensing \cite{DEApaper}. Nevertheless, DEAs are operated under high electric fields, making them prone to electric breakdowns and electrical aging. There are versions of DEAs, namely Fault-tolerant DEAs, that demonstrate continued operation after dielectric failure, but their strains are usually compromised and are found below 5\%. Furthermore, DEAs are difficult to scale up and deliver high forces, thus not ideal for developing a gripper. 
    \subsubsection{Hydraulically Amplified Self-healing Electrostatic Actuators.} In light of the above, HASEL actuators seem to combine the best of both worlds, the versatility of the fluidic actuators and the muscle-like performance and self-sensing properties of the DEAs. Because the dielectric in HASELs is a liquid, it enables the actuator to self-heal itself immediately after an electric breakdown. Even multiple breakdowns won't influence the functionality, assuming the polymer film doesn't take any damage. Besides that, HASEL actuators have also proven to show fast actuation speeds and low power consumption, all while being lightweight \cite{peanohasel}.
     
\subsection{HASEL Challenges}
    The main challenge when working with HASELs is finding a sweet spot in the force/strain relation of the actuator under a certain operating voltage $V$. Because the force is proportional to the voltage squared as derived from the analytical model of a Peano-HASEL presented in \cite{haselcharacteristicl}, 
    \begin{equation} 
       F = \frac{w}{4t}\frac{cos(\alpha)}{1-cos(\alpha)}\epsilon_{0}\epsilon_{r}V^{2}
       \label{forcevoltage}
    \end{equation}    
    \noindent it is generally desired to operate at the highest voltage possible without causing any electrical breakdowns. On the other hand, when developing a system meant for drone integration that interacts with its environment, it's typically better to stay within a lower voltage region while still ensuring sufficient strength and strain. Although the voltage has the biggest influence on the HASEL's performance, as seen in Equation \ref{forcevoltage}, the pouch geometry and the shell material also play a significant role \cite{haselcharacteristicl}. In particular, parameters like the width $w$, the thickness $t$ or the dielectric permittivity of the material $\varepsilon_{r}$ are of importance. On that note, for strain improvements, the initial pouch length $L_{p}$ also becomes relevant. Because Maxwell pressure is independent of the electrode area, the actuation force and strain can be scaled by adjusting the ratio of electrode area to the total surface of the elastomeric shell. That means an actuator with larger electrodes displaces more liquid dielectric, generating a larger strain but a smaller force because the resulting hydraulic pressure acts over a smaller area. Conversely, an actuator with smaller electrodes displaces less liquid dielectric, generating less strain but more force because the resulting hydraulic pressure acts across a larger area \cite{donutpapbase}.
\subsection{RAPTOR Quadcopter}\label{RAPTOR_quadcopter}
\begin{figure}[htp]
        \centering
        \includegraphics[width=0.7\columnwidth] {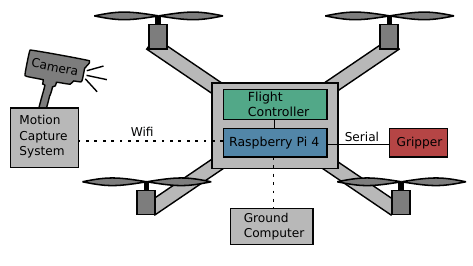} 
        \caption{The quadcopter uses a motion capture system for position feedback control. A Raspberry Pi 4 is used as an onboard computer. The trajectory generation runs on a ground station computer which sends both position and gripper commands to the onboard computer. There, the commands are forwarded to the flight controller and the gripper over a serial connection.}
        \label{RAPTOR_System}
    \end{figure}

    RAPTOR is an acronym that stands for Rapid Aerial Pick-and-Transfer of Objects by Robots. The system consists of a conventional quadcopter frame where a gripper can be attached to the bottom. \cite{Appius2022RAPTORRA} deployed a soft gripper to pick objects with different geometries in a rapid swooping motion. The system architecture, seen in Figure \ref{RAPTOR_System} allows for an effortless integration of various gripper types.
\subsection{Actuator Manufacturing}
\begin{figure}[htp]
        \centering
        \includegraphics[width=0.9\columnwidth] {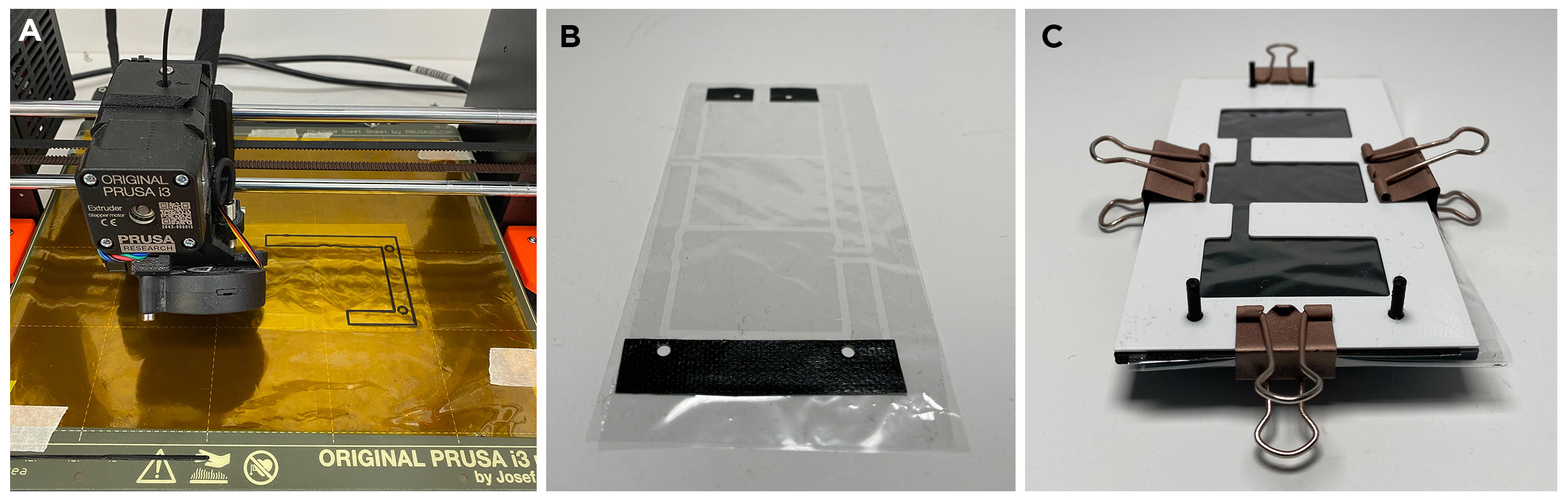} 
        \caption{Manufacturing process of a HASEL actuator. A) Sealing two Mylar sheets to obtain the desired actuator geometry using a Prusa 3D printer. B) Sealed polymer film with reinforced mounting holes. The holes primarily served the fixation during the airbrushing process. C) Actuator clamped for airbrushing.}
        \label{Manuf}
    \end{figure}

 The stencil and sealing pattern of the actuator were designed in CAD. Once all the components were available, the polyester film was sealed. It turned out that using Mylar 850, a co-extruded, one-side heat sealable polyester film, worked best. Other materials like Hostaphan were also used throughout the many iterations, though resulting in significantly weaker sealing lines while also requiring higher temperatures. For the sealing process, we used a Prusa 3D printer with a custom configuration to provide the optimal sealing temperature at an adequate speed. After that, we reinforced the holes of the polymer film with structural tape and added the electrodes using an airbrush. The described process is depicted in Figure \ref{Manuf}. 
\noindent The final steps included filling the pouches using the Envirotemp FR3 Fluid as dielectric and attaching the leads to the electrodes with a two-component epoxy adhesive. We decided to forego a commonly used electrode channel to make the actuator more compact. Regarding the structural part of the gripper, we chose to 3D print all of the parts using PLA. Furthermore, we came up with a sandwich configuration to make the finger more sturdy compared to previous work \cite{spider}. We used standard packing tape for the hinge layer that connects all the bottom pieces and guides the structure during rotation. To join the top with the bottom plates, we decided on simple tape to omit the need for screws to save weight and space. Lastly, we utilized the same adhesive for mounting the HASEL actuator to the backplate structure. 
\section{Inverse VS. Classical Gripper Approach}
\begin{figure}[htp]
        \centering
        \includegraphics[width=\columnwidth] {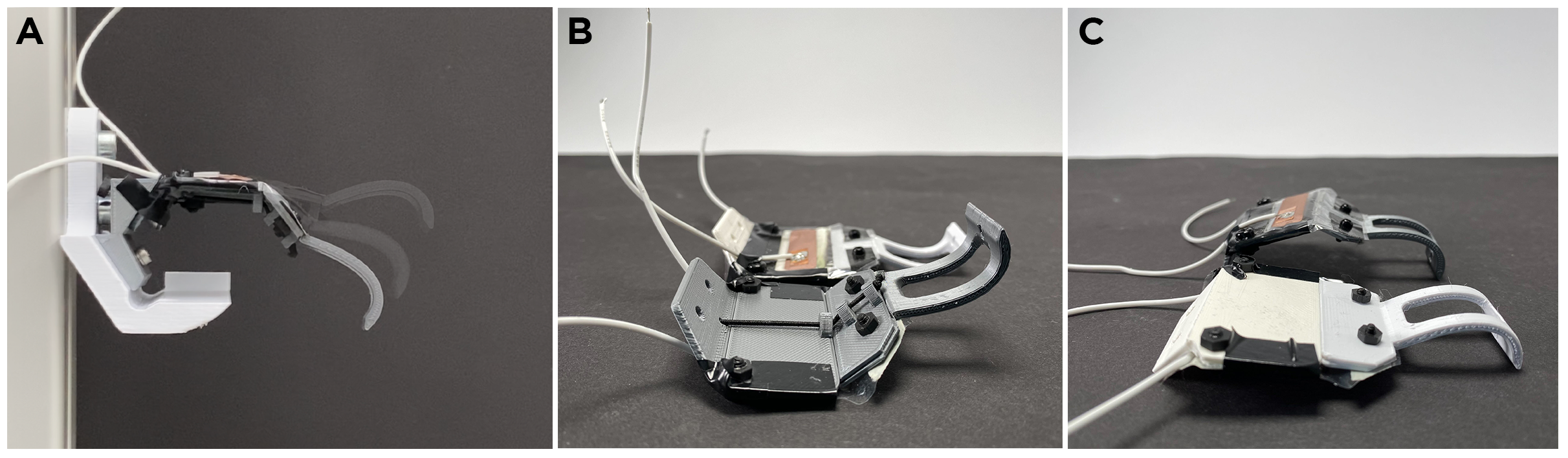} 
        \caption{Side-by-side comparison of a traditional and inverse gripper. A) Reverse gripper with a pouch length of 40 mm and width of 6 mm. In contrast to a classic gripper, the actuator is mounted on the outside of the toe. On the inside, a 1 mm thick cord was used for prestressing the system in the order of 0.2 N. B) \& C) Visual comparison of the reverse gripper with a traditional model. The black tape is used to insulate the gripper and to avoid arching at high voltages.}
        \label{RevGripperPlusComparison}
    \end{figure}

    One of the most notable aspects of an eagle catching its prey is that it doesn't fly around with its toes constantly open. Eagles only stretch their talons right before they grab their prey. Thus, we considered designing a prestressed system that would open upon actuation but is closed by default. That is essentially the inverse process of a classic gripper that contracts during operation and the reason we decided to call this concept inverse gripper. We initially wanted to prestretch our inverse gripper prototype with a spring. Since there are numerous springs on the market and we were only interested in the functionality at first, we decided to use a stretchable cord of 1 mm thickness. When loaded, the selected cord performed very similarly to our traditional one-hinge actuator designs. Figure \ref{RevGripperPlusComparison} shows the inverse gripper and its side-by-side comparison to a classic gripper design. 
    \noindent Because of the increased bearing surface through the prestress of the cord and the facilitated pulling due to the initial displacement, the inverse gripper showed a 14.2\% greater maximum force output when compared to a traditional design. More specifically, a model from the PWT series was selected to match the geometry of the inverse gripper. Having the actuator mounted on the outside of the gripper makes it more prone to failure. Because of that and simplicity, we decided to pursue a regular gripper design despite slightly weaker performance.
\section{Actuator Types and Configurations}
    Instead of building a complete gripper with every iteration, which would have been time-consuming, we followed a bottom-up approach. That means we first focused on one hinge, then determined the optimal configuration of one toe and finally built the entire system. The baseline for our joint optimization was a HASEL actuator with a pouch geometry of 40x10x10 mm, denoting the electrode's length and width as well as the pouch width, respectively. Those values originated from a preliminary design study we conducted to find proper starting-point dimensions. Since we aimed to design a system for an aerial vehicle, we intended to save as much weight and space as possible while not significantly compromising the deflection or force output. Consequently, the length and width of the pouch played a significant role. Thus, we divided the single-hinge optimization into two test cases:
    \begin{enumerate}[label=\Roman*.]
    	\item Pouch Width Test (PWT)
    	\item Pouch Length Test (PLT)
    \end{enumerate}
\subsection{Pouch Width Test (PWT)}
\begin{figure}[htp]
        \centering
        \includegraphics[width=\columnwidth] {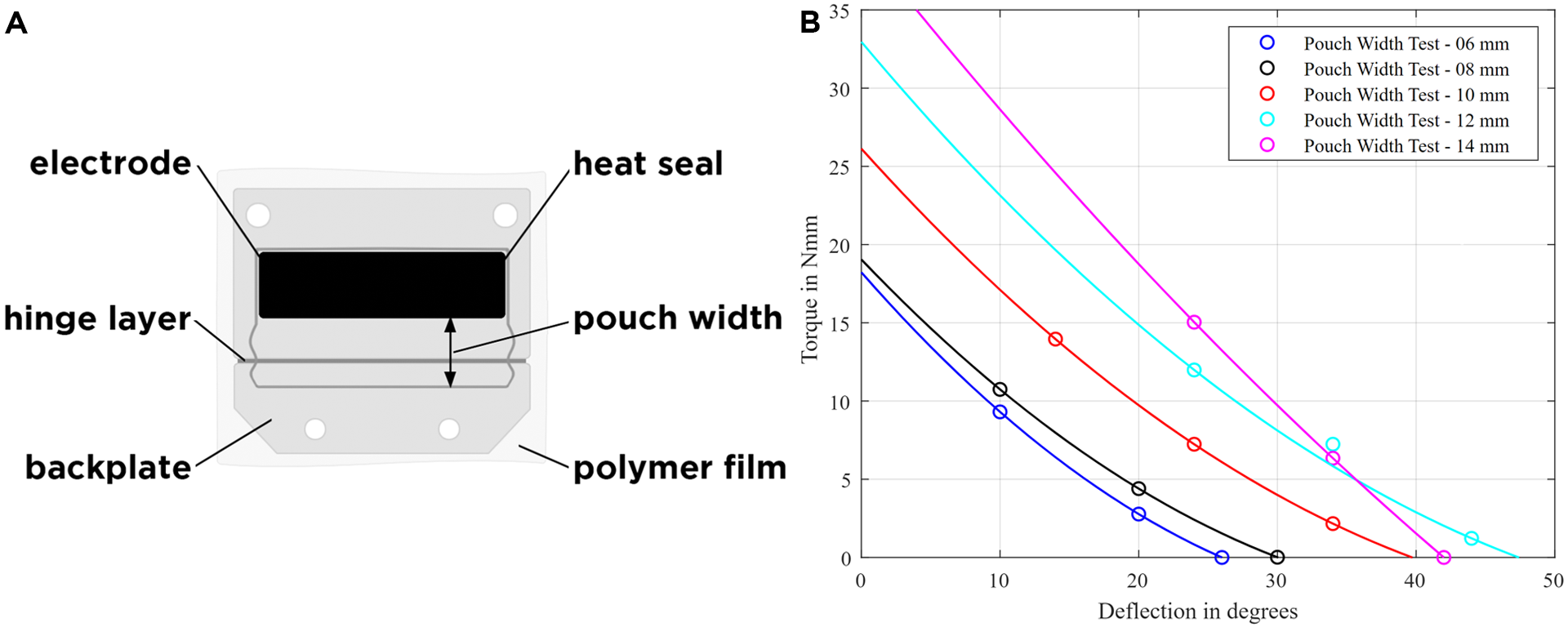} 
        \caption{Performance comparison of PWT actuators. A) Illustration of a PWT actuator. B) Exceeded torque from the individual actuator types at different deflection angles. Second-order curve fits approximate the measurement response.}
        \label{PWTANDIllustration}
    \end{figure}

  In the PWT series, the primary attention went towards the maximum deflection the actuator can achieve. As shown in \cite{haselcharacteristicl}, increasing the deflection of a Peano-HASEL is not a linear process. Instead, the free strain reaches a pinnacle before declining again. Therefore, we investigated how that observation translates from a linear to an angular HASEL actuator. We defined pouch widths ranging from 6 mm to 14 mm for our experiments. If we take the relationship between the width of the electrode (L$_{e}$) and the width of the whole pouch (L$_{p}$), we end up with the following range for our test case:
    \begin{equation}
        0.417 \leq f_{e, PWT} \leq 0.625 \quad \text{and} \quad f_{e} = \frac{L_{e}}{L_{p}}
    \end{equation}
    \noindent While the measured torque kept rising by increasing the pouch width, the actuator reached a maximum deflection of nearly 50° at a 12 mm pouch width. That equals an electrode coverage of $f_{e,PWT 12} = 0.455$. After that, the highest possible deflection of the talon decreased, as shown in Figure \ref{PWTANDIllustration}.
\subsection{Pouch Length Test (PLT)}
    Conversely, in the PLT series, the achievable deflection should not change by varying the electrode length. What does change is the force output. Tapering the electrodes when stacking multiple joints or actuator types would make for a more realistic toe design. Hence, we investigated how narrowing the electrode length impacted the gripping force. As predicted by equation \ref{forcevoltage}, the torque and pouch length share a linear relation. Comparing the torque values from an actuator with a \SI{40}{\milli\meter} to one with a \SI{30}{\milli\meter} electrode length yields a loss of roughly 40\%, which notably impairs the functionality of the gripper. As a result, tapering the electrodes when stacking multiple hinges or actuation types was not an option.
\subsection{Scorpion Concept Test}
    As discussed in Section \ref{related work}, a famous example of an early bio-mimicking planar HASEL was the scorpion tail concept. Hence, when mentioning "Scorpion" throughout this paper, we are referring to the design of having a base electrode with pouches stacked adjacently. The Scorpion is exciting because large deflections can be achieved while saving a lot of space due to the absence of electrodes between the pouches. For our Scorpion concept, we chose to further develop the initial design by combining it with the assets of the spider-inspired joints \cite{spider}. Eventually, we decided on a base electrode with a height of 20 mm, instead of the 10 mm used so far, to avoid any large pre-deflections. Subsequently, we used two adjacent pouches, each with a 10 mm width. 

\begin{figure}[htp]
        \centering
        \includegraphics[width=0.6\columnwidth] {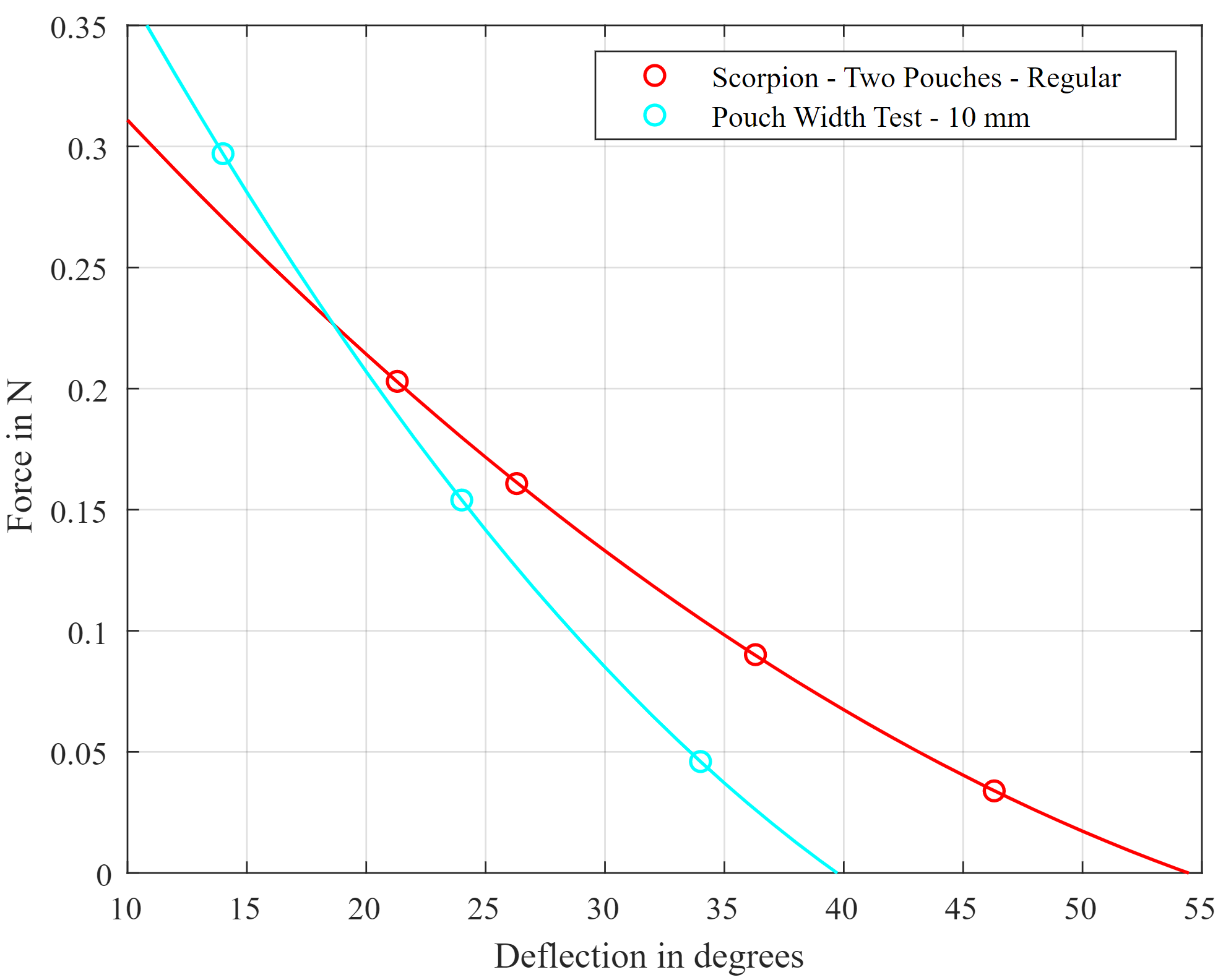} 
        \caption{Force output comparison between a two-pouch Scorpion actuator and a 10 mm pouch width PWT actuator. The PWT actuator performs better until reaching a deflection of around 19°. Subsequently, the scorpion demonstrates a higher force output combined with a deflection increase of about 30\%.}
        \label{ScorpionVSPWT}
    \end{figure}

    Comparing Scorpion grippers with different liquid dielectrics, the one with Envirotemp FR3 showed the best force output and deflection, with the runner-up containing silicone oil. Moreover, the preliminary design study revealed that the dielectric struggles to prevail through the channel connecting the two pouches if its width was chosen below 4 mm. Thus, a 5 mm wide channel was selected for our Scorpion prototype. Undergoing several actuation tests, the Scorpion prototype demonstrated a deflection of more than 50°. Also, the force output was satisfying with 0.2 N under 20° deflection. A side-by-side comparison of the performance between the Scorpion and the single-hinge concept with similar dimensions ($f_{e,Scorpion} = f_{e,PWT 10}$) demonstrated that the single-hinge actuator is superior until a deflection of around 19°. After that point, the two-pouch-design delivered notably higher forces and contracted almost 30\% more, as depicted in Figure \ref{ScorpionVSPWT}.
%
\subsection{Complete Finger Configuration} \label{ConfigGripper}
\begin{figure}[htp]
        \centering
        \includegraphics[width=\columnwidth] {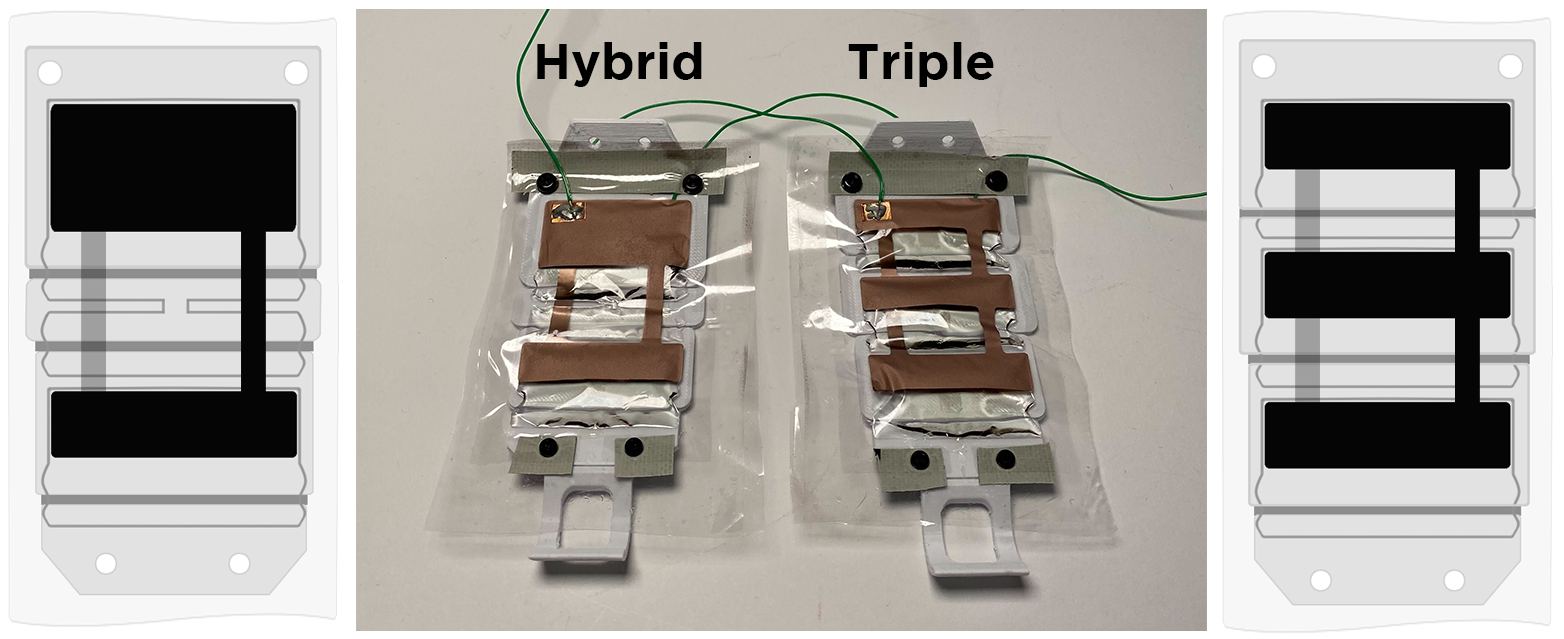} 
        \caption{Side-by-side comparison between the Hybrid and Triple concept. The illustrations and the picture in the middle visualize the idea and prototype of the chosen configurations, respectively. The dimensions for the Hybrid were 20/10 mm electrode width, 10/10/12 mm pouch width, and for the Triple, 10/10/10 mm electrode width and 12/12/12 mm pouch width. All values are in order, from base to tip.}
        \label{HybTrip_System_Illustration}
    \end{figure}

\begin{figure}[htp]
        \centering
        \includegraphics[width=\columnwidth] {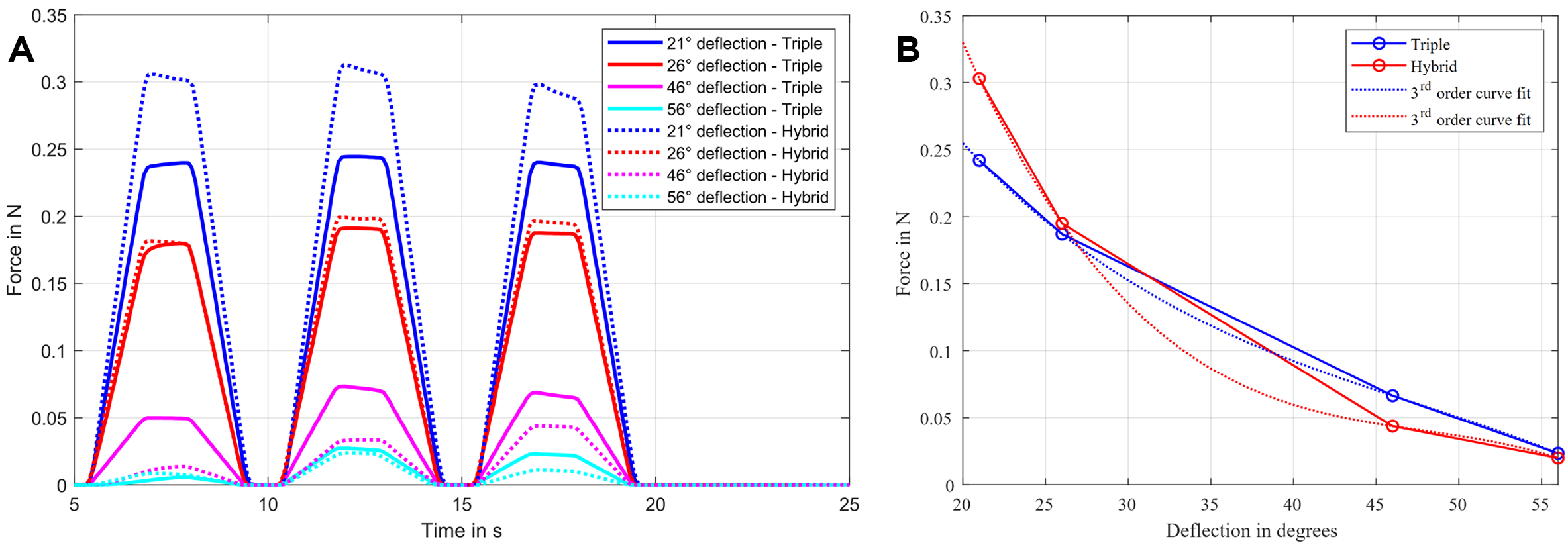} 
        \caption{Performance comparison between the Hybrid and Triple configuration. The electrodes of both actuators were made using Electrodag 502. A) Raw data comparison between the two proposed toe designs, indicating a notable advantage of the Hybrid. B) Evolution of the actuator's exhibited force based on different angles. The Hybrid performs better under small deflections, while the Triple seems to have the upper hand at medium to large contractions.}
        \label{HybVSTrip}
    \end{figure}

\begin{figure}[htp]
        \centering
        \includegraphics[width=0.8\columnwidth] {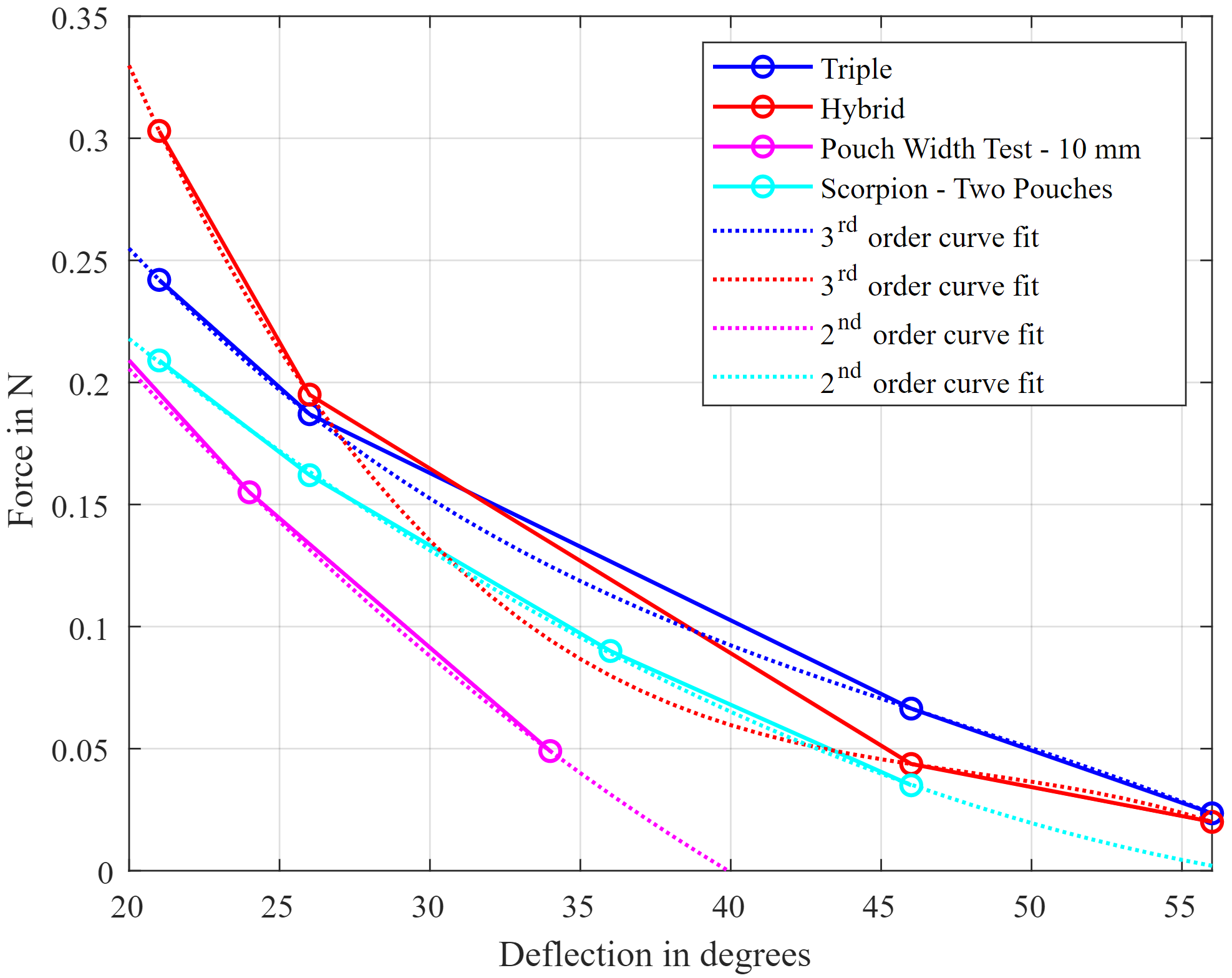} 
        \caption{Side-by-side comparison between the single-hinge, Scorpion, Hybrid and Triple design. The graph provides a comparative overview of the actuators' performance relative to each other.}
        \label{FullComparison}
    \end{figure}

    After we determined the optimal pouch geometry for our actuator and compared the two main actuation types, we looked into different ways of combining everything to a complete toe configuration. We found that the two gripper designs described below and illustrated in Figure \ref{HybTrip_System_Illustration} both offer great performance while being as compact as possible.
    \begin{enumerate}[label=\Roman*.]
        \item In Figure \ref{ScorpionVSPWT}, the classical single-hinge design showed a force output of 0.2 N under a deflection of 20°. Connecting three of those in series should yield a deflection of at least 80° while still generating a force of roughly 0.1 N, assuming a linear stacking behavior. Because of its similarity to a tripe-pouch Peano-HASEL, we abbreviated this concept to "Triple". 
    	\item Because the Scorpion design demonstrated an exemplary deflection behavior, we fused it with a classical single-hinge design to further improve the deflection and increase the force output at the tip of the talon. We chose to name that said concept "Hybrid".
    \end{enumerate}
Switching the electrode material from BP-B311 to Electrodag 502, a carbon paint commonly utilized in similar publications, led to an overall force increase of up to 26.5\%. After that, we compared the performance of the Triple with the Hybrid concept, as depicted in Figure \ref{HybVSTrip}. It turned out that for slight deflections up to 26°, the Hybrid displays a higher force output. Subsequently, the Triple holds an advantage over the Hybrid until they reach a contraction of about 56°, where they meet again with a tiny force output of around 0.025 N. 
    Putting the two concepts into perspective with the classical single-hinge and Scorpion actuator gave rise to Figure \ref{FullComparison}. It can be observed that the advantage of the Hybrid at lower deflections stems from the additional single-hinge in the front. Since the Scorpion part of the Hybrid takes care of the main deflection, the front actuator only displays minimally, nearly exceeding its full force potential. Looking at the performance of a single-hinge and comparing it with the output from the Triple implies a linear stacking behavior. Thus, explaining the consistency throughout actuation by sharing the deflection more or less equally among all three pouches. 
\section{Drone Integration}
\begin{figure}[htp]
        \centering
        \includegraphics[width=\columnwidth] {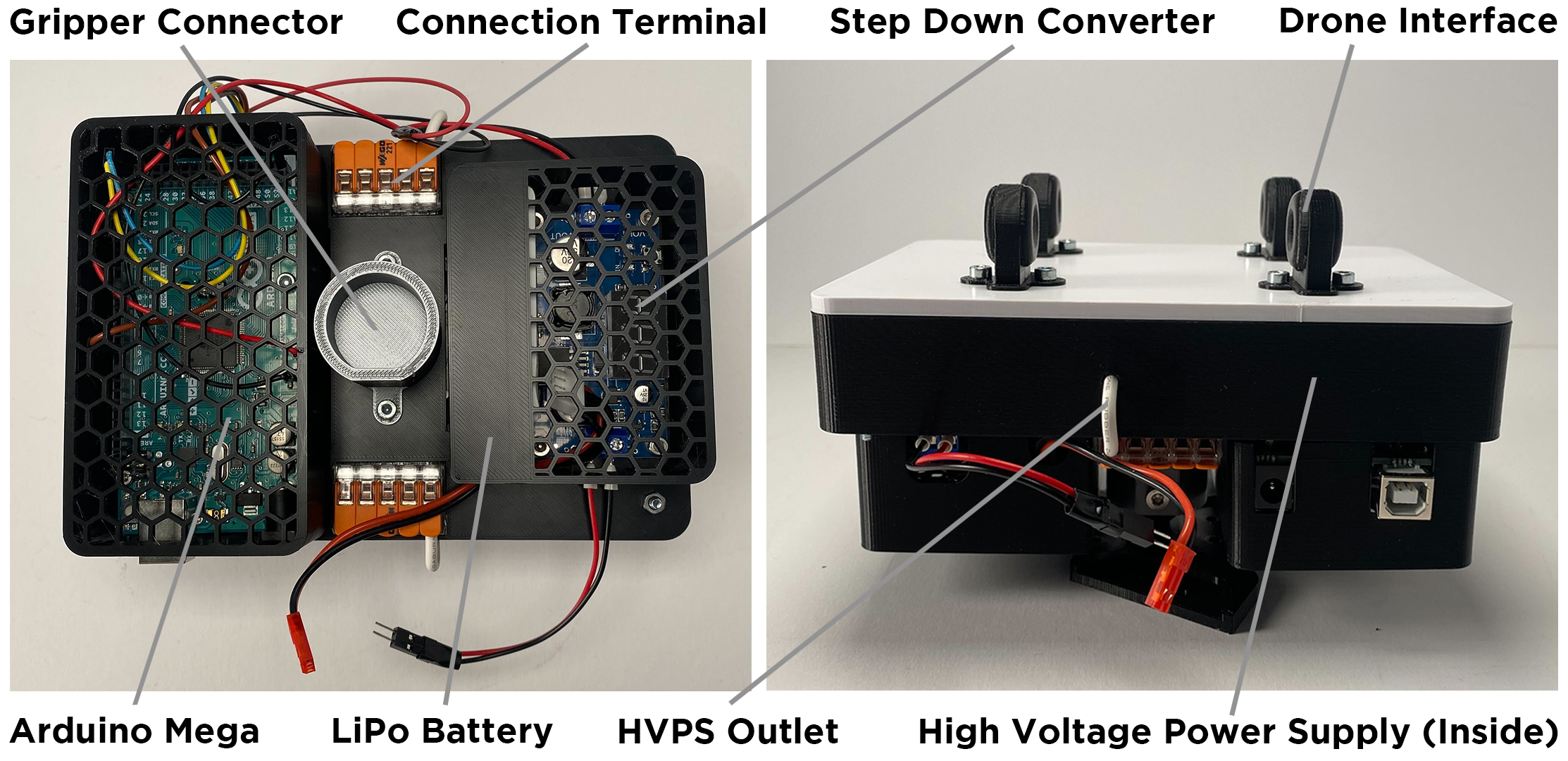}
        \caption{Untethered power supply and control system of the gripper. The gripper operates using an Arduino Mega. A 3S LiPo battery with a step-down module powers the Arduino and the high voltage power supply that provides the gripper with the necessary voltage.}
        \label{Box_Desc}
    \end{figure}

  After identifying the best gripper configuration for our case, we designed the electronics to supply the system with the necessary voltage of up to 10 kV. As seen in figure \ref{Box_Desc}, the system is composed of a high voltage power supply (HVPS) \cite{haselimproved} on the insight of the main box, a micro-controller (Arduino Mega), a step-down converter (to supply the HVPS and Arduino with 5 V) and a 3S 11.1 V / 500 mAh LiPo battery, as seen in Figure \ref{Box_Desc}. The electronic components are covered by self-designed snap-fit cases for quick and easy accessibility while offering protection and clean cable management. The connection terminals distribute the high voltage and ground path via cables leading to the gripper. The gripper is mounted in the middle by inserting the counterpart into the connector.
    \subsection{Flight Experiments}
\begin{figure}[htp]
        \centering
        \includegraphics[width=\columnwidth] {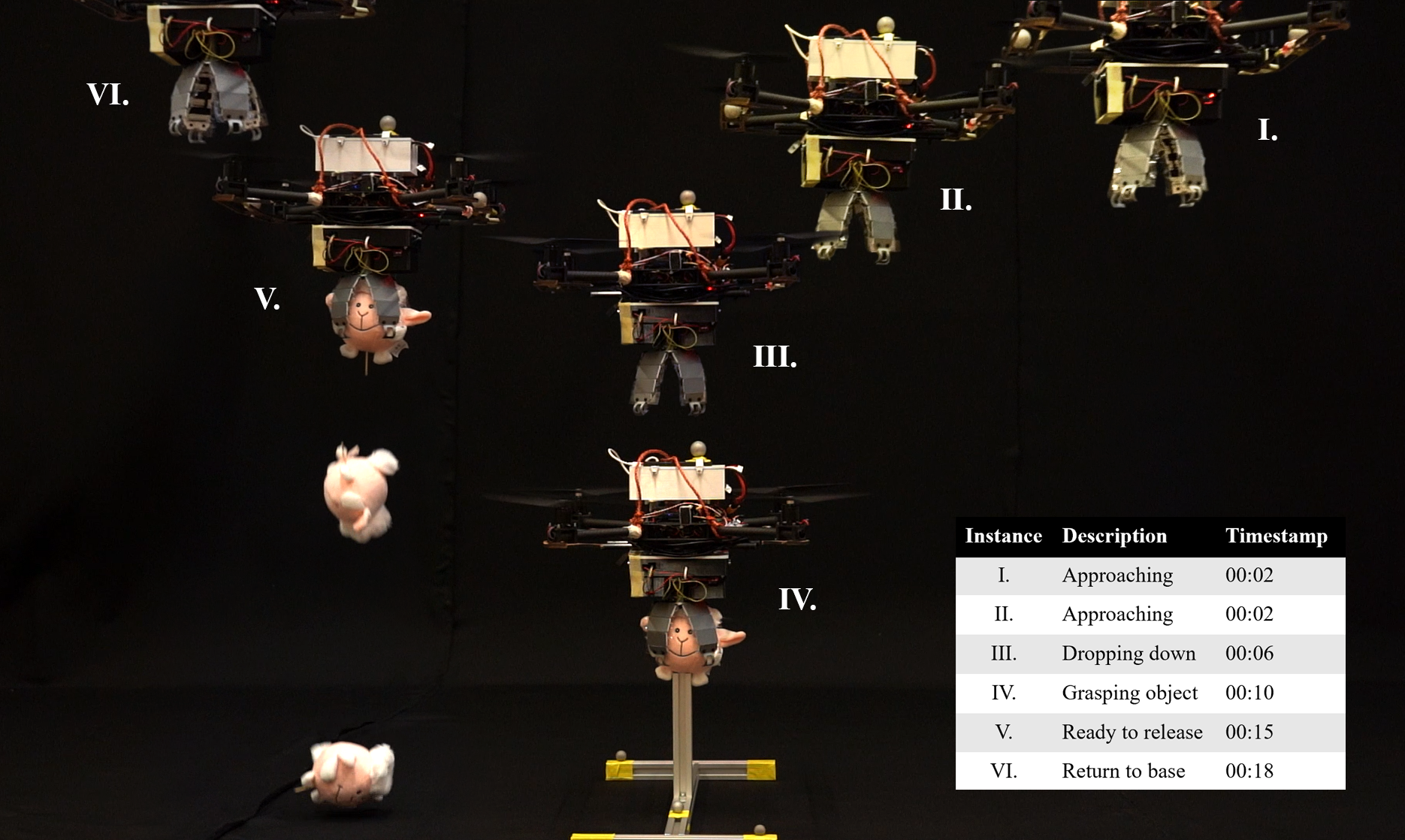} 
        \caption{Entire system demonstrating aerial object manipulation capabilities. The drone picks and places an object weighing 76 g with a gripper operating voltage of \SI{8}{\kilo\volt}.}
        \label{GripMotion_3}
    \end{figure}

    Finally, we combined our gripper system with the RAPTOR quadcopter. For safety reasons, we separated the drone's power supply from the one of the gripper in case of any failure like an electrical breakdown. A successful aerial grasp, as in Figure \ref{GripMotion_3}, requires time-critical actuation, which was achieved by using the existing RAPTOR system architecture described in section \ref{RAPTOR_System}. By installing a serial connection between the onboard computer and the micro-controller on the gripper, commands could be delivered at the right time. Once the serial actuation signal was detected, the microcontroller sent a pulse width modulation (PWM) to the gripper. The mechanical connection between the drone and the gripper was accomplished by inserting two carbon rods that passed through small connectors on both subsystems.
    \noindent As a test object, we used a stuffed animal a bit larger than the size of a fist weighing 76 g. We had to fixate the plush with a small stick to prevent it from being blown away by the downwash caused by the quadcopter propellers. A major difference that had to be taken into account was that the actuation time to fully contract the toes took slightly longer using the untethered HVPS than the TREK 20/20C-HS used in the static ground tests.
    \begin{figure}[htp]
        \centering
        \includegraphics[width=\columnwidth] {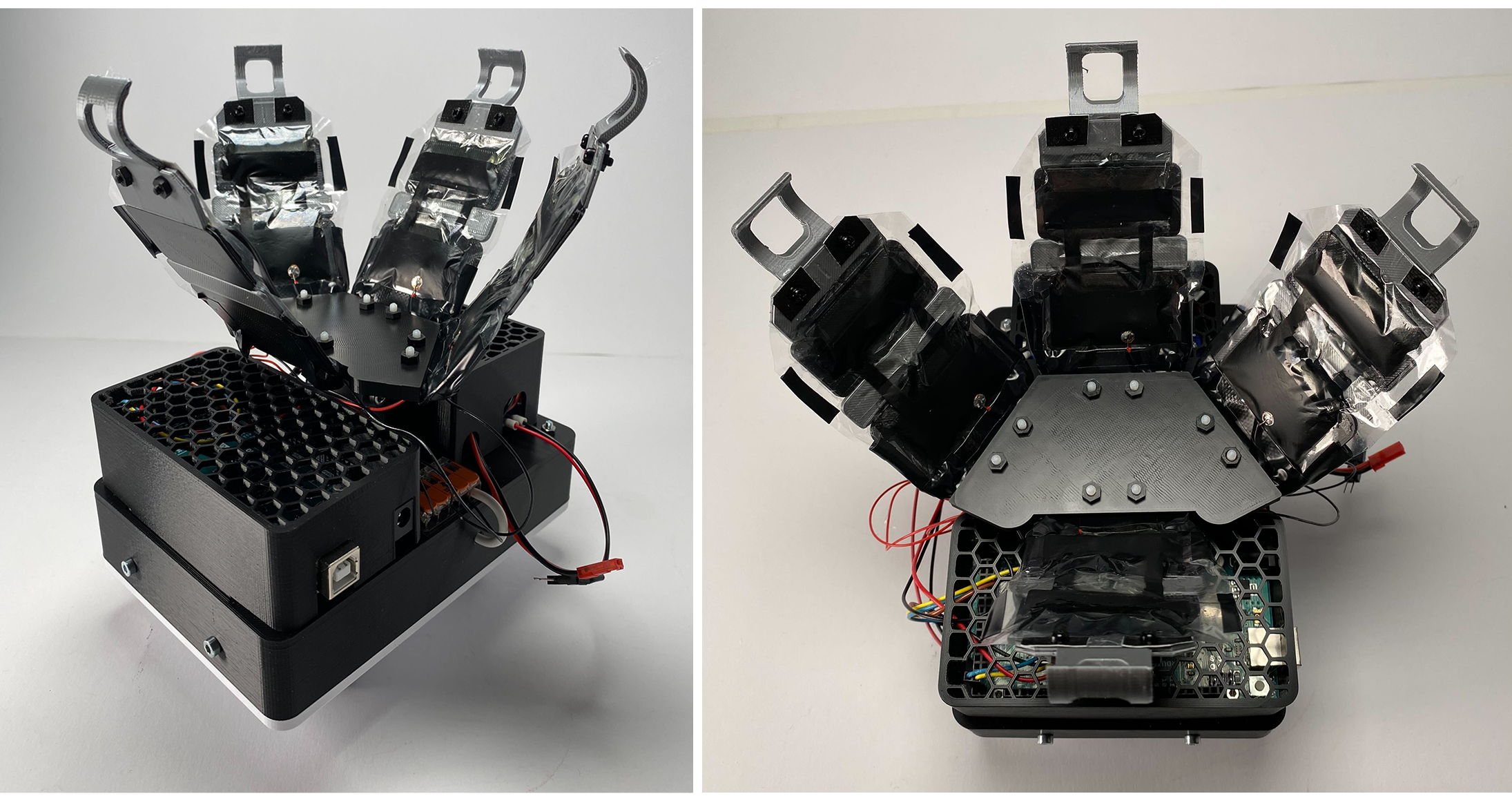} 
        \caption{Full gripper assembly with the eagle actuator configuartion. The system is shown from a side and top view, respectively.}
        \label{FullSysGrip}
    \end{figure}
\section{Conclusion and Future Work}
    We proposed a bio-inspired HASEL gripper which we integrated as part of a bigger system for the first time. We investigated various pouch geometries and configurations leading us to a satisfying trade-off between force output and deflection. We showed that by combining different actuator types into a hybrid design, the actuator could achieve a higher force output under small deflections compared to the traditional Triple concept. Figure \ref{FullSysGrip} shows the eagle-inspired gripper. As opposed to our original idea, the biomimetic characteristics were less predominant in the final design than expected. Through early testing, it became clear that narrowing down the area of rotation to mimic an eagle's talon actuation will not yield a sufficient force output. Since our system did not require perching capabilities, we departed from the hallux (back toe) in later versions. Departing from the hallux yielded an equally spaced arrangement of the four toes, evident in \ref{GripMotion_3}, which enabled us to better employ their individual force contributions and improve redundancy. Stationary experiments and in-flight tests confirmed the gripper's performance of being able to pick and place objects weighing up to 76 g without the use of additional material to increase friction. Compared to previous HASEL grippers, our hinges proved to be significantly more sturdy and durable, even enduring minor crashes. In addition, we did not experience any electrical breakdowns during several in-flight tests, even at operating voltages of up to \SI{10}{\kilo\volt}. In the future, weight and space savings can be made by redesigning the HVPS and selecting a different micro-controller. Ideally, a micro-controller that's powerful enough to keep the supplied voltage constant during operation.
%
\bibliographystyle{ieeetr}

\bibliography{references} 

\end{document}